\documentclass{article}
\usepackage{ijcai23}

\usepackage{times}
\usepackage{soul}
\usepackage{url}
\usepackage{adjustbox}
\usepackage{array}
\usepackage[hidelinks]{hyperref}
\usepackage[utf8]{inputenc}
\usepackage[small]{caption}
\usepackage{graphicx}
\usepackage{amsmath}
\usepackage{amsthm}
\usepackage{booktabs}
\usepackage{algorithm}
\usepackage{algorithmic}
\usepackage[switch]{lineno}

\title{Instruct-FinGPT: Financial Sentiment Analysis by Instruction Tuning of General-Purpose Large Language Models}

\author{
    Boyu Zhang$^{2*}$, Hongyang (Bruce) Yang$^1$\thanks{Equal contribution.}, Xiao-Yang Liu$^1$\thanks{Corresponding author.}
    \affiliations
    $^1$Columbia University; $^2$Swinburne University of Technology  
    \emails
    \{HY2500, XL2427\}@columbia.edu; boyu.zhang68@gmail.com
}

\begin{document}

\maketitle

\begin{abstract}
% Sentiment analysis is indispensable in deciphering financial articles, news, and social media, yielding essential insights into market movements. However, conventional models are often hindered by limitations such as insensitivity to numbers and difficulty in discerning ambiguous sentiment absent of context. In this paper, we explore the potential of Large Language Models (LLMs) in financial sentiment analysis. LLMs, trained on a vast amount of data and encompassing a considerable number of parameters, have the potential to surmount existing obstacles. Motivated by the inherent general knowledge and reasoning capabilities of LLMs, we adopt a novel approach, \textit{instruction fine-tuning}, which leverages the extensive text generation and analysis capabilities. This approach aims to improve sentiment analysis performance with minimal fine-tuning data, opening up new avenues for research in the finance sector. Experimental results demonstrate that the proposed \textit{Instruct-FinGPT} outperforms the FinBERT model in sentiment classification, numerical sensitivity, and contextual understanding.
Sentiment analysis is a vital tool for uncovering insights from financial articles, news, and social media, shaping our understanding of market movements. Despite the impressive capabilities of large language models (LLMs) in financial natural language processing (NLP), they still struggle with accurately interpreting numerical values and grasping financial context, limiting their effectiveness in predicting financial sentiment. In this paper, we introduce a simple yet effective instruction tuning approach to address these issues. By transforming a small portion of supervised financial sentiment analysis data into instruction data and fine-tuning a general-purpose LLM with this method, we achieve remarkable advancements in financial sentiment analysis. In the experiment, our approach outperforms state-of-the-art supervised sentiment analysis models, as well as widely used LLMs like ChatGPT and LLaMAs, particularly in scenarios where numerical understanding and contextual comprehension are vital.
\end{abstract}

\section{Introduction}

Financial sentiment analysis, the task of discerning investor sentiment from financial articles, news, and social media, is an essential instrument for comprehending and forecasting market movements. Conventional models often struggle with several difficulties, including insensitivity to numeric values, difficulties interpreting sentiment without explicit context, and the challenges associated with financial jargon, multi-lingual data, temporal dependency, insufficient labeled data, and the inherent noise in social media data.

Large Language Models (LLMs) have been pivotal in mitigating some of these challenges, demonstrating a significant contribution to the field of financial natural language processing (NLP). One distinguishing feature is LLMs' inherent general knowledge garnered during pre-training on vast and diverse corpora, including financial texts. However, LLMs do not have enough financial context. Their performance in interpreting numerical values is often inadequate and may struggle to accurately determine sentiment when the context is absent or ambiguous. These challenges underline the need for improved models that can adeptly understand the intricate nuances of financial sentiment analysis.

In response to these challenges, our study explores the potential of instruction tuning of general-purpose LLMs for sentiment analysis in the finance sector. 
In this study, we investigate two primary research questions: \textit{\textbf{1) How to enable LLMs to address the issue of numerical sensitivity in financial sentiment analysis? and 2) What is the role of contextual understanding in improving financial sentiment analysis? }}

We propose \textit{Instruct-FinGPT} by instruction tuning \cite{wei2021finetuned} a pre-trained LLM (namely LLaMA \cite{touvron2023llama}). Through this approach, we transform the classification based sentiment analysis dataset into a generation task, thereby allowing LLMs to apply their extensive training and superior analytical capabilities more effectively. The ultimate goal of Instruct-FinGPT is to enhance the performance in financial sentiment analysis by minimizing the requirement of fine-tuning data and maximizing the contextual understanding and numerical sensitivity inherent to LLMs. By introducing this novel method, we aspire to push the boundaries of current methodologies, opening up promising avenues for future exploration in the realm of financial sentiment analysis.

The primary contributions of this paper are as follows:
%大语言模型具备world knowledge和数字推理能力，我们是想在实验里面分析证明这一点
%如何分析：case study，把FPB里面数据集跟数字有关的，不包含明确的incrase和decrease的句子跳出来，然后跟FinBERT

\begin{itemize}
% \item We study a novel application of LLMs, financial sentiment analysis.
% \item We adopt an instruction tuning method that transforms sentiment analysis into a generation task, maximizing the model's learning effectiveness.
\item We design an instruction-tuned FinGPT model for financial sentiment analysis. This model surpasses both general-purpose LLMs and state-of-the-art supervised models in benchmark performance, despite utilizing only a small amount of instruction data and training resources.
\item We address the critical issue of numerical sensitivity in financial sentiment analysis, a component often neglected by existing models, enhancing the model's ability to accurately interpret sentiment from financial news.
\item We underscore the importance of contextual understanding in financial sentiment analysis, leveraging the inherent general knowledge of LLMs for improved performance in sentiment analysis, especially when the context is missing or vague.
\end{itemize}

Our study provides new insights into the application of LLMs for financial sentiment analysis, offering potential solutions to some of the enduring challenges in the field.

\section{Related Work}

The task of sentiment analysis, particularly in the financial domain, has been a significant area of research in the field of Natural Language Processing (NLP). There are several works in literature \cite{xing2018natural,loughran2011liability,tai2013automatic,hamilton2016inducing,day2016deep,chan2017sentiment,sohangir2018big,araci2019FinBERT,mishev2020evaluation} that utilize different methodologies for performing financial sentiment analysis, ranging from lexicon-based techniques to machine learning and deep learning approaches.

One noteworthy work is by Araci \cite{atkins2018financial}, which presents an ensemble of traditional machine learning algorithms for predicting the direction of stock market movement based on financial news articles. While this work made strides in using machine learning for financial sentiment analysis, it does not extensively address the challenges related to numerical sensitivity or contextual understanding.

In terms of deep learning approaches, the transformer-based model BERT \cite{devlin2018bert} has been widely used for sentiment analysis tasks due to its powerful context understanding capability. However, BERT and its derivatives typically require substantial amounts of labeled data for fine-tuning, which might be challenging to obtain in the financial domain.

More recently, FinBERT \cite{araci2019FinBERT}, a variant of BERT designed explicitly for the financial domain, was developed to address these issues. FinBERT has been fine-tuned on the financial text and has shown promising results in financial sentiment analysis. Nonetheless, it suffers from limitations such as insensitivity to numerical values and struggles with the context where the necessary information may be missing. FLANG \cite{shah-etal-2022-flang} additionally presents financial assessment benchmarks across five distinct NLP tasks within the financial sector, along with the incorporation of conventional benchmarks prevalent in prior research. 

%While BloombergGPT \cite{wu2023bloomberggpt} has shown impressive performance in sentiment analysis tasks, it should be noted that it is a proprietary model developed by Bloomberg. The model benefits from their exclusive access to specialized financial data, enabling them to train a financial LLM (FinLLM).

While BloombergGPT \cite{wu2023bloomberggpt} demonstrates impressive performance in sentiment analysis tasks, there are inherent challenges to its accessibility and applicability for broader usage. The model, proprietary to Bloomberg, was trained on a vast corpus of specialized financial data, which may not be readily available to others. Moreover, the resources required to train such a model are substantial (1.3M GPU hours, a cost of around \$5M). This is in contrast to our approach, which demonstrates substantial effectiveness with a significantly smaller corpus and less computational resources (estimated around less than \$300 per training), making it more feasible for wider deployment.

Our work stands distinct in its focus on leveraging the power of LLMs, their inherent general knowledge, and reasoning capabilities to perform sentiment analysis in the financial domain. We explore a novel instruction tuning approach and demonstrate its effectiveness in our experiments.

%%%%%%%%%%%%%%%%%%%%%%%%%%%%%%%%%%%%%%%%%%%%%%%%%%%%%%%%%%
\section{Our Method}
%  core: 
% The core of our methodology lies in leveraging the potential of LLMs for financial sentiment analysis. We aim to exploit the inherent advantages of these models, including their vast contextual understanding and sensitivity towards numerical data, to outperform traditional models like FinBERT. 
Despite the pre-trained LLMs such as GPT-3 and LLaMA can acquire the general abilities for solving various tasks, increasing studies have shown that LLM’s abilities can be further adapted according to specific goals. Our approach uses instruction tuning to adapt the general-purpose LLMs to financial sentiment analysis, enhancing their understanding of numerical values and context in this specific task. The process involves transforming the sentiment analysis task from a classification task to a text generation task, which aligns better with the capabilities of LLMs. Further, we use the transformed dataset to instruction finetune the LLMs in a supervised learning way. Last, we map the generated outputs into sentiment labels during inference.

\subsection{Instruction Tuning}

%2.1 formulate sentiment analysis dataset into instruction finetuning dataset\\
%2.2 supervised finetuning the llama-7b via deepspeed-chat (results in 10 min)\\
%2.3 mapping the generated outputs into sentiment labels.

We adopt the instruction tuning method of an LLM on financial sentiment analysis datasets. This process is divided into three main steps:

\subsubsection{Formatting Financial Sentiment Analysis Dataset into Instruction Tuning Dataset}

\begin{figure*}
\centering
\includegraphics[scale = 0.363]{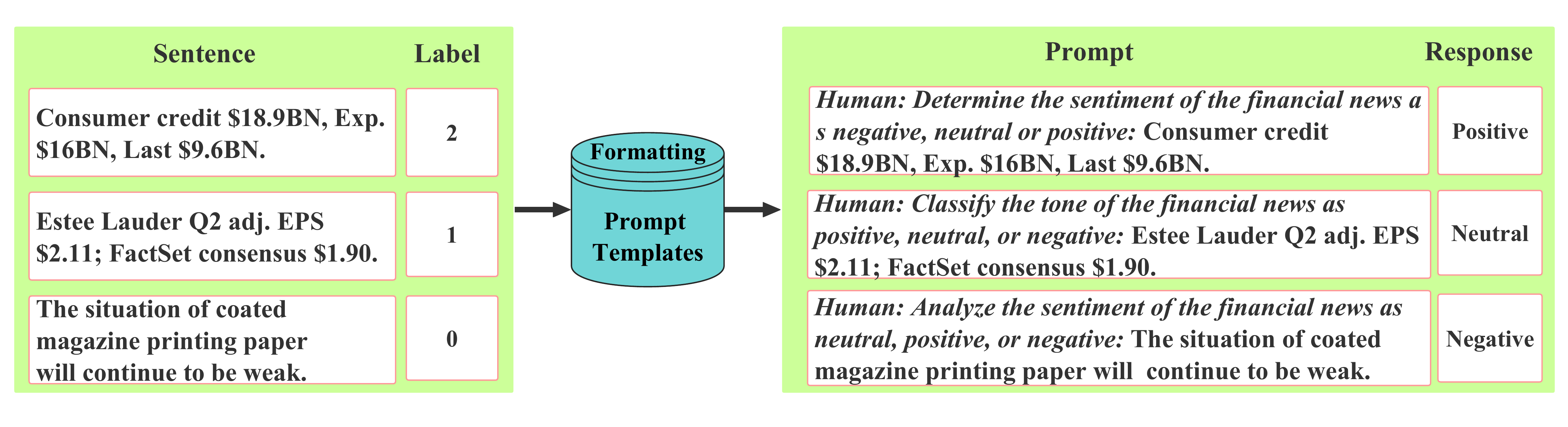}
\caption{Formatting sentiment analysis dataset into instruction tuning dataset.}
\label{fig:framework}
\vspace{-1mm}
\end{figure*}

The existing financial sentiment analysis datasets are formatted as text classification task where the \textbf{inputs} are the financial news or headlines and the \textbf{outputs} are integer-type labels representing \textit{positive}, \textit{negative} and \textit{neutral} sentiments. Our first step is to formulate these classification datasets into instruction-formatted dataset.

Following \cite{zhao2023survey}, we create 10 human-written \textbf{instructions} describing the task of financial sentiment analysis, and formulate each sample from the original dataset by combining one randomly selected \textbf{instruction} with the \textbf{input} and \textbf{output} in the format of "Human: [\textbf{instruction}] + [\textbf{input}], Assistant: [\textbf{output}]". This process is shown in Fig \ref{fig:framework}. %In this way,  the model is instructed to generate whether the sentiment is positive, neutral, or negative.
%prompt 10个每次randomly选一个防止overfit

\subsubsection{Instruction Tuning LLaMA-7B}
%流程图来一个

% We conduct instruction tuning on LLaMA-7B specifically designed for the financial sentiment analysis task. It also provides the flexibility of single model finetuning for domain-specific tasks such as ours.

% Instruction tuning refers to the process of fine-tuning pre-trained language models (LLMs) using a set of formatted instances in natural language \cite{wei2021finetuned}. It is closely connected to supervised fine-tuning. During training, we utilize these formatted instances to fine-tune LLMs through supervised learning, such as training with a sequence-to-sequence loss. 
While pretrained LLMs possess capabilities such as reasoning, understanding numbers, world knowledge, and multilingualism, they struggle to effectively apply these abilities to specific tasks. This limitation hinders their ability to achieve state-of-the-art (SOTA) performance on specific tasks, thus restricting their application potential. For instance, \cite{wei2021finetuned} found that the zero-shot performance of LLMs is significantly lower compared to their few-shot performance. In our scenario, we leverage instruction data, which typically includes numeric values, financial context, and financial jargon, to provide supervised signals. Through instruction tuning, we align the LLM's capabilities with the sentiment analysis labels, achieving a more precise and nuanced understanding of sentiments expressed in financial texts which enables it to outperform both pretrained LLMs and supervised models specifically designed for financial sentiment analysis.

We illustrate our approach using instruction tuning with the LLM model called LLaMA-7B as an example to validate our ideas. Instruction tuning involves fine-tuning pre-trained LLMs by leveraging a collection of formatted instances in natural language \cite{wei2021finetuned}. It is a method closely aligned with supervised fine-tuning. During the training process, we specifically employ the formatted instances to fine-tune the LLaMA-7B LLM using a supervised learning approach, i.e., training with a sequence-to-sequence loss. This choice allows us to showcase the effectiveness and applicability of instruction tuning in enhancing the financial sentiment analysis performance of LLMs like LLaMA-7B.

\subsubsection{Mapping the Generated Outputs into Sentiment Labels}
% Post fine-tuning, we map the model-generated text into sentiment labels. The generation task output (positive, neutral, negative) is interpreted as sentiment labels. The capacity of the LLM is thus effectively channeled towards the accurate determination of financial sentiment.

% Our instruction tuning approach applied to LLMs addresses the limitations observed in existing models, including FinBERT. It improves the accuracy of financial sentiment analysis, better leveraging the contextual understanding and numerical sensitivity of the model.

Since the instruction finetuned LLaMA-7B is an autoregressive generative model, even though we train it using instruction templates to guide its output towards the desired sentiment judgments, it still has the possibility of generating free-style text. Therefore, we need to map the model's output back to the specified three emotions for proper evaluation. Our approach is as follows: if the model's output contains "positive," "negative," or "neutral" terms, we map it to the corresponding label; otherwise, we consider it as "neutral" sentiment.

%\subsection{Comparison Between Instruct Tuned LLMs and General-Purpose LLMs for Sentiment Analysis}
\subsection{Comparison Between LLMs and FinBERT for Sentiment Analysis}
% 1. large-scale pretraining context of llms lead to more comprehensive world knowledge. (comparision between the scale of )\\
% 2. numerical awareness. (refers to some reports or articles)\\
% 3. decoder-only vs encoder
Our approach employs LLMs and compares their efficacy in sentiment analysis with the well-established FinBERT model. The comparison is based on three pivotal aspects:
\begin{itemize}
\item \textbf{Contextual understanding}: LLMs have an advantage due to their large-scale pretraining on diverse data. This provides them with a more comprehensive general knowledge, enabling a superior understanding of the context compared to FinBERT. The diversity and richness of the training datasets of LLMs are unmatched, providing them with a well-rounded knowledge that outshines FinBERT's capability.

\item \textbf{Numerical sensitivity}: Financial texts often incorporate significant numerical data, which plays a crucial role in conveying the sentiment. LLMs, with their inherent numerical sensitivity, exhibit an enhanced capacity for interpreting the sentiment implied by numerical fluctuations. Refer to certain scholarly reports for in-depth studies on this characteristic of LLMs.
\item \textbf{Decoder-only vs encoder-only models}: FinBERT is an encoder-only model which encodes the input sequence into a representation and relies on a separate classifier to make predictions based on the encoded representation. On the other hand, the employed LLM is a decoder-only model which can generate the entire output sequence, including the class label, directly from a latent representation or fixed-length vector. This character allows the LLMs easily adapt to various tasks without modifying the model structure while the encoder-only models require the development of task-specific classifiers, which can be more labor-intensive.

% While FinBERT is an encoder-only model, designed to understand the input and generate a subsequent output, LLMs like GPT-3 are decoder-only models. This means they generate the output word by word, considering the entire context, providing an edge in tasks like text generation and, by extension, sentiment analysis. 

% a decoder-only model generates the entire output sequence, including the class label, directly from a latent representation or fixed-length vector. On the other hand, an encoder-only model encodes the input sequence into a representation and relies on a separate classifier to make predictions based on the encoded representation.
\end{itemize}

\section{Performance Evaluation}

% \begin{table*}[htb]
% \centering
% \begin{tabular}{|c|c|c|c|}
% \hline
% \textbf{Datasets}&\multicolumn{3}{|c|}{\textbf{Models}} \\
% \cline{2-4} 
%  & \textbf{FinBERT} & \textbf{Ours-7B}  & \textbf{Ours}
% \\ \hline
% % \textbf{FPB (ACC)} &1e-4  &1e-4  &1e-4  
% % \\ \hline
% % \textbf{FPB (F1)} &32 &32     &32     
% % \\ \hline
% % \textbf{FiQA (ACC)}  & 10& 10& 10
% % \\ \hline
% % \textbf{FiQA (F1)} & 1 & 1 & 1      
% % \\ \hline
% \textbf{Twitter Val (ACC)} & 0.73 & 0.79 & 1       
% \\ \hline
% \textbf{Twitter Val (F1)} & 0.6682 & 1 & 1      
% \\ \hline
% \textbf{Numerical (ACC)} &0&0   &0         
% \\ \hline
% \textbf{Numerical (F1)} &0&0   &0         
% \\ \hline
% \textbf{Contextual (ACC)} &0&0   &0         
% \\ \hline
% \textbf{Contextual (F1)} &0&0   &0         
% \\ \hline
% % \textbf{Max Token Length} &512 &512&512
% % \\ \hline
% \textbf{Training Time} &14mins &58mins&108mins
% \\ \hline
% \end{tabular}
% \caption{Zero-Shot Experimental Results on the Twitter News dataset}
% \label{tab:parameters}
% \end{table*}

\begin{table*}
    \centering
        \begin{tabular}{cccccc}
            \toprule
            \multicolumn{3}{c}{\textbf{Datasets}} & \multicolumn{3}{c}{\textbf{Models}} \\
             \midrule
            \textbf{Name}&\textbf{Size}&\textbf{Metrics}&\textbf{FinBERT} & \textbf{LLaMA-7B} & \textbf{Instruct-FinGPT-7B} \\
            \midrule
            Twitter Val    & 2388 & Acc & 0.725   & 0.54  & \textbf{0.880}      \\
                           &      & F1 & 0.668   & 0.36  & \textbf{0.841}       \\
            \midrule
           Testing Time &     &    &18 seconds (1 GPU) & 498 seconds (8 GPUs)& 498 seconds (8 GPUs)  \\
            \midrule %[2pt]  
            %\midrule %[2pt]  
            Numerical      & 117  & Acc & 0.633   & 0.60  & \textbf{0.837}      \\
                           &     & F1 & 0.630   & 0.42  & \textbf{0.795}      \\
            \midrule %[2pt]  
            Contextual     & 20  & Acc & 0.50   & 0.55  & \textbf{0.80}     \\
                           &     & F1 & 0.22   & 0.34  & \textbf{0.63}      \\
            \bottomrule

        \end{tabular}
    \caption{Experimental results on the Twitter financial news sentiment validation, numerical, and contextual datasets}
    \label{tab:booktabs}
\end{table*}

% We set up experiments to validate our hypotheses, conducting sentiment analysis tasks on financial datasets with our instruction fine-tuned FinGPT model. The effectiveness of the model is measured against the benchmark performance of FinBERT.

In this section, we evaluate the effectiveness of our proposed method from three perspectives: general sentiment analysis, numerical understanding, and general knowledge supplementing. To validate our method's performance, we compare it against state-of-the-art sentiment analysis model, FinBERT, and the general-purpose LLM, ChatGPT.

Our experimental results validate the effectiveness of our approach. With only a small amount of fine-tuning data, our model consistently achieves superior performance in sentiment analysis compared to FinBERT and ChatGPT. %These promising results demonstrate the effectiveness of utilizing the pre-training general knowledge and reasoning capabilities of large models.

\subsection{Datasets}
%分一下数据集
Our training data is an amalgamation of the Twitter Financial News dataset \cite{twitter2022finance} and FiQA dataset \cite{fiqa}, resulting in a comprehensive collection of $10,501$ samples.

\subsubsection{Training Datasets}

\begin{itemize}
    \item \textbf{Twitter financial news sentiment training}: This dataset is a corpus of news tweets that pertain to the financial sector and is exclusively in English. Its primary purpose is the classification of financial sentiment within the context of Twitter discussions. The dataset comprises 9,540 samples for training, each annotated with one of three labels: Bearish, Bullish, or Neutral.
    \item \textbf{FiQA dataset}: This dataset, which is readily accessible via HuggingFace, includes 961 samples. Each sample has been annotated with one of three labels: positive, neutral, or negative, denoting the sentiment conveyed in the corresponding text. 
\end{itemize}

\subsubsection{Testing Datasets}

\begin{itemize}
    \item \textbf{Twitter financial news sentiment validation (Twitter Val)}: This dataset, accessible through HuggingFace, contains 2,390 samples annotated with three labels: Bearish, Bullish, or Neutral.
    \item \textbf{Numerical sensitivity dataset (numerical)}: This dataset, which we automatically filtered from Twitter Val, includes 117 samples. These samples contain at least two numerical values related to financial indicators without strong indication words such as 'raise', 'fall', 'increase', 'decrease'. 
    \item \textbf{Contextual understanding dataset (contextual)}: This dataset, which we randomly selected from Twitter Val, includes 20 samples. These samples lack the essential contexts to make a sentiment prediciton. 
    \item \textbf{Financial PhraseBank (FPB) dataset}: This dataset \cite{malo2014good} comprises 4,840 samples randomly extracted from financial news articles available on the LexisNexis database. The samples were carefully annotated by a team of 16 annotators with backgrounds in finance and business, ensuring high quality annotations.

\end{itemize}

\subsection{Model Training}

% \paragraph{Instruct-FinGPT-7B}
% The training parameters are listed in Table \ref{tab:parameters}. Our Instruct-FinGPT-7B model is instruction fine-tuned for $10$ epochs. This is done using the AdamW optimizer \cite{loshchilov2017fixing}, with a batch size of $32$, an initial learning rate of $1e-5$, a weight decay of 0.1. We constrain the maximum length of input texts to 512 tokens. The fine-tuning of our Instruct-FinGPT-7B model is carried out on 8 A100 (40GB) GPUs, which took $58$ minutes for training.

The training parameters are given in Table \ref{tab:parameters}. For our Instruct-FinGPT-7B model, we initialize it with LLaMA-7B model and perform instruction tuning over $10$ epochs. The training process utilizes the AdamW optimizer \cite{loshchilov2017fixing}, with a batch size of $32$, an initial learning rate of $1e^{-5}$, and a weight decay of $0.1$. To maintain efficiency, we set a maximum input text length of $512$ tokens. We utilize DeepSpeed \cite{deepspeed} for the fine-tuning process on 8 A100 (40GB) GPUs, resulting in a total training time of $58$ minutes.

\begin{table}[h]
\centering
\begin{tabular}{ll}
\hline
Parameter        & Value           \\ \hline
Learning rate    & 1e-5            \\
Weight Decay     & 0.1           \\
Batch size       & 32              \\
Training epochs  & 10              \\
LR Scheduler     & CosineAnnealing \\
Num warmup Steps & 0               \\
Max Token Length & 512             \\ 
GPUs             & 8 A100 (40GB)               \\

\hline
\end{tabular}
\caption{Training parameters.}
\label{tab:parameters}
\end{table}

\begin{table*}[htbp]
\centering
\begin{adjustbox}{valign=m}
\begin{tabular}{|p{2in}|c|c|c|c|c|}
\hline
\textbf{News} & \textbf{True Value} & \textbf{FinBERT} & \textbf{ChatGPT 3.5} & \textbf{ChatGPT 4.0} & \textbf{Instruct-FinGPT}
\\ \hline
Pre-tax loss totaled euro 0.3 million, compared to a loss of euro 
2.2 million in the first quarter of 2005. & \textbf{Positive} & Negative & Negative & \textbf{Positive} & \textbf{Positive} 
\\ \hline
Madison Square Garden Q2 EPS \$3.93 vs. \$3.42. & \textbf{Positive} & Negative & \textbf{Positive}  & \textbf{Positive}  & \textbf{Positive}   
\\ \hline
Consumer credit \$18.9BN, Exp. \$16BN, Last \$9.6BN. & \textbf{Positive} & Neutral & \textbf{Positive}  & \textbf{Positive}  & \textbf{Positive}   
\\ \hline
 Estee Lauder Q2 adj. EPS \$2.11; FactSet consensus \$1.90. & \textbf{Neutral} & \textbf{Neutral} & Positive & Positive & \textbf{Neutral}  
\\ \hline
% \textbf{Training Time} &14mins &58mins&100mins
% \\ \hline
\end{tabular}
  \end{adjustbox}
\caption{Examples and results on the numerical sensitivity dataset.}
\label{tab:numerical}
\end{table*}

\begin{table*}[t]
\centering
\begin{adjustbox}{valign=m}
\begin{tabular}{|p{2in}|c|c|c|c|c|}
\hline
\textbf{News} & \textbf{True Value} & \textbf{FinBERT} & \textbf{ChatGPT-3.5} & \textbf{ChatGPT-4.0} & \textbf{Instruct-FinGPT}
\\ \hline
The situation of coated magazine printing paper will continue to be weak. & \textbf{Negative} & Neutral & \textbf{Negative}  & \textbf{Negative}  & \textbf{Negative} 
\\ \hline
Boeing announces additional order for 737 MAX planes. & \textbf{Neutral} & Positive & Positive  & Positive  & Positive
\\ \hline
Boeing: Deliveries 24 Jets in November. & \textbf{Positive} & Neutral & \textbf{Positive}  & Neutral  & \textbf{Positive} \\  
\hline
 PPD's stock indicated in early going to open at \$30, or 11\% above \$27 IPO price. & \textbf{Neutral} & Positive & Positive & Positive & \textbf{Neutral} \\
\hline
\end{tabular}
  \end{adjustbox}
\caption{Examples and results on the contextual understanding dataset.}
\label{tab:contextual}
\end{table*}

\subsection{Baseline Models}

\paragraph{LLaMA-7B} \cite{touvron2023llama} We obtained the LLaMA-7B\footnote{We use LLaMA-7B for research and education purposes.} model from Meta and use it for inference, keeping the same inference setting as our Instruct-FinGPT-7B.

\paragraph{FinBERT} We obtained the FinBERT model from the Hugging Face Model Hub. The FinBERT model is used for sentiment analysis after pre-processing raw data, which includes tokenizing the text and padding or truncating it to fit the model's max input length. Once pre-processed, the data is run through FinBERT for inference, providing sentiment analysis results (positive, negative, or neutral) for each text input.

\paragraph{ChatGPT}

The utilization of OpenAI's ChatGPT API for sentiment analysis comprises a streamlined four-step process:
\begin{enumerate}
    \item \textbf{API setup}: This involves setting up the OpenAI Python client, which serves as an interface to interact with the ChatGPT API.
    \item \textbf{Data preparation}: The Instruction Tuning dataset as shown in Figs. \ref{fig:framework} is employed for the inference with the ChatGPT model.
    \item \textbf{API call}: Due to existing limitations, the GPT-3.5 API is used for requests. The GPT-4.0 version is currently unavailable for programmatic access and can only be interacted with via a web interface.
    \item \textbf{Response interpretation}: The response from the API includes the sentiment of the text directly. This direct sentiment output simplifies the task of sentiment analysis.
\end{enumerate}

\subsection{Evaluation and Analysis}

To evaluate the performance of our model, we test it on a benchmark financial sentiment analysis dataset and contrast the results with those of FinBERT. The key evaluation metrics center around the model's capability to manage numerical values and comprehend sentiment within various contexts.

\paragraph{Performance Metrics}

The primary performance metrics for our model include accuracy, and F1-score. Accuracy measures the proportion of correct predictions, 
% precision assesses the proportion of true positive predictions, recall (or sensitivity) measures the ability of the model to find all relevant instances, 
and the F1-score represents the harmonic mean of precision and recall.

\paragraph{Overall Performance}

Based on the evaluation results in Table \ref{tab:booktabs}, our instruction tuned LLaMA-7B (Instruct-FinGPT-7B) consistently outperforms both FinBERT and LLaMA-7B across all three datasets in terms of accuracy and F1 score. Especially, comparing our Instruct-FinGPT-7B with the original LLaMA-7B model (without instruction tuning), it is evident that the instruction tuning method significantly improves the model's performance on financial sentiment analysis. 

\paragraph{Analysis of Numerical Sensitivity}

Numerical data plays a crucial role in financial sentiment analysis, as it often reflects important financial indicators. In Table \ref{tab:numerical}, we assess the models' ability to comprehend and interpret sentiment associated with numbers.

\begin{itemize}
    \item Example 1: This is an example from FinBERT, where FinBERT failed in this case. However, ChatGPT 4.0 and Instruct-FinGPT correctly recognize the substantial decrease in the loss from 2.2 million to 0.3 million, indicating a positive sentiment.
    \item Example 2: The increase in EPS is correctly identified as a positive sentiment by all models except FinBERT.
    \item Example 3: The exceeding of consumer credit expectations and the previous value is recognized as a positive sentiment by all models except FinBERT.
    \item Example 4: The statement about Estee Lauder and FactSet consensus is neutral, as it merely states the facts without indicating a positive or negative sentiment.
\end{itemize}

Our model demonstrates varying levels of effectiveness in understanding and interpreting sentiment associated with numerical data.

\paragraph{Analysis of Contextual Understanding}

The ability of our model to interpret sentiment in different contexts is an important aspect of its performance evaluation. Financial news can be nuanced, and a statement that may appear negative in one context could be neutral or even positive in another. We assess the models' performance in contextual understanding based on the examples provided in Table \ref{tab:contextual}.

\begin{itemize}
    \item Example 1: This is an example from FinBERT, where FinBERT failed in this case. But ChatGPT and Instruct-FinGPT recognized that the situation of coated magazine printing paper is expected to remain weak, indicating a negative outlook for the industry. LLMs' language understanding capabilities and knowledge of financial contexts enable them to accurately interpret such statements and predict the sentiment.
    \item Example 2: In this specific case, indicating that Boeing has received more orders for their aircraft. It looks like positive news. However, without further context, it's challenging to determine the sentiment accurately. The lack of specific details about the order, the customer, or any potential implications can make it difficult to assess the sentiment correctly.
    \item Example 3: The sentiment of the financial news is positive. The statement highlights that Boeing delivered 24 jets in November, indicating a successful and productive month for the company.
    \item Example 4: All of the models failed on this one. The opening price of a stock is higher than the IPO price doesn't necessarily indicate the stock is rising from its current market price.
\end{itemize}

Overall, our model demonstrates a better understanding of the contextual sentiment in these examples compared to FinBERT and ChatGPT. It successfully recognizes the negative sentiment in Example 1 and accurately identifies the neutral sentiment in Example 2 and the positive sentiment in Example 3. These results highlight the importance of contextual understanding in financial sentiment analysis and the variations in performance across different models.

\begin{table}[t]
   \centering
    \begin{tabular}{ccccc}% 其中，tabular是表格内容的环境；c表示centering，即文本格式居中；c的个数代表列的个数
        \toprule %[2pt]设置线宽     
          Performance & ChatGPT 3.5 & LLaMA-7B & Ours-7B
          \\ %换行
        \midrule %[2pt]  
         FPB (ACC) & 0.64 & 0.60 &  \textbf{0.76} 
         \\
         FPB (F1) & 0.51 & 0.40 & \textbf{0.74}
         \\
         %FiQA SA (ACC)& \textbf{75.07}  & 51.60 & 53.12  \\
        %FiQA SA (F1)& \textbf{75.07}  & 51.60 & 53.12  \\

        \midrule %[2pt]
        %Estimated Cost & \$3.2 M & \$4.22 M & \$11.26 M &  \\
        %\bottomrule %[2pt]     
    \end{tabular}
       \caption{Zero-shot evaluation between ChatGPT and Instruct-FinGPT on the entire dataset of financial phaseBank.}
   \label{table:zero-shot}
\end{table}

\paragraph{Zero-Shot Generalization to Other Financial Datasets}

Finally, we evaluate the zero-shot ability of our model, which refers to how well the model can generalize to other unseen financial datasets. A model with strong zero-shot capabilities can provide more robust and versatile results in real-world applications. We compare our Instruct-FinGPT with ChatGPT3.5 and LLaMA-7B on the full FPB dataset. Here we do not compare with FinBERT because it uses FPB as the training set. 

The evaluation results are shown in Table \ref{table:zero-shot}. Based on these results, it can be concluded that the instruction tuned LLaMA-7B model performs the best among the three, achieving the highest accuracy and F1 score. The fine-tuning process with sentiment instruction data seems to have improved the model's ability to capture sentiment in financial phrases, resulting in better zero-shot performance compared to both ChatGPT and the original LLaMA-7B model.
% These evaluation measures together provide a comprehensive understanding of the model's capabilities and effectiveness in the field of financial sentiment analysis. The results are then compared to the performance of FinBERT, providing a comparison of our model's capabilities to existing models in the field.

% \section{Results}

% The results demonstrate that our LLM significantly outperforms FinBERT in terms of numerical sensitivity and context understanding. Moreover, our model shows strong generalizability across different types of financial texts and languages.

\section{Conclusion and Future Work}

In this paper, we have presented an innovative approach for financial sentiment analysis by harnessing the general knowledge and reasoning capabilities of LLMs. Our method represents a substantial contribution to the field of sentiment analysis, demonstrating that instruction tuning of an LLM can yield superior performance with a small amount of task-specific data. Our findings pave the way for future research into the potential of LLMs for a broad range of financial tasks.

\textbf{Disclaimer: We are sharing codes for academic purposes under the MIT education license. Nothing herein is financial advice, and NOT a recommendation to trade real money. Please use common sense and always first consult a professional before trading or investing.}

\bibliographystyle{named}
\bibliography{ijcai23}

\end{document}